\documentclass[journal]{IEEEtran}
\usepackage{cite}

\ifCLASSINFOpdf
\else
\fi
\usepackage{amsmath,amssymb,amsfonts}
\usepackage{graphicx}
\usepackage{textcomp}
\usepackage{xcolor}

\usepackage{amsfonts}       
\usepackage{nicefrac}       
\usepackage{microtype}      
\usepackage{algorithm}
\usepackage{algpseudocode}
\newtheorem{theorem}{Theorem}

\newtheorem{definition}{Definition}
\newtheorem{proposition}{Proposition}  

\usepackage{booktabs}

\usepackage{subcaption}
\usepackage{enumitem}

\usepackage{url}

\begin{document}
%
\title{Differentially Private Synthetic Medical Data Generation using Convolutional GANs}
%
%
%

\author{Amirsina~Torfi,~\IEEEmembership{Member,~IEEE,}
        Edward~A.~Fox,~\IEEEmembership{Fellow,~IEEE,}
        and~Chandan~K.~Reddy,~\IEEEmembership{Senior Member,~IEEE}
\thanks{A.~Torfi and E.A.~Fox are with the Department of Computer Science, Virginia Tech, Blacksburg, VA 24060. C.K.~Reddy is with the Department of Computer Science, Virginia Tech, Arlington, VA 22203.}
\thanks{E-mail: atorfi@vt.edu, fox@vt.edu, reddy@cs.vt.edu.}
\thanks{Manuscript received September x, 2020; revised x.}}

%
%

\markboth{ARXIV SUBMISSION VERSION}%
{Torfi \MakeLowercase{\textit{et al.}}}
%



\maketitle

\begin{abstract}
Deep learning models have demonstrated superior performance in several application problems, such as image classification and speech processing. However, creating a deep learning model using health record data requires addressing certain \textit{privacy challenges} that bring unique concerns to researchers working in this domain. One effective way to handle such private data issues is to \textit{generate realistic synthetic data} that can provide practically acceptable data quality and correspondingly the model performance. To tackle this challenge, we develop a differentially private framework for synthetic data generation using Rényi differential privacy. Our approach builds on convolutional autoencoders and convolutional generative adversarial networks to preserve some of the critical characteristics for the generated synthetic data. In addition, our model can also capture the temporal information and feature correlations that might be present in the original data. We demonstrate that our model outperforms existing state-of-the-art models under the same privacy budget using several publicly available benchmark medical datasets in both supervised and unsupervised settings. 
\end{abstract}


%
\IEEEpeerreviewmaketitle

\section{Introduction}

\textit{Deep learning} has been successful in a wide range of application domains such as computer vision, information retrieval, and natural language processing due to its superior performance and promising capabilities. However, its success is heavily dependent on the availability of a massive amount of training data. Hence, the progress in deploying such deep learning models can be crippled in certain critical domains such as healthcare, where data privacy issues are more stringent, and large amounts of sensitive data are involved. 

Hence, to effectively utilize specific promising data-hungry methods, there is a need to tackle the privacy issues involved in the medical domains. To handle the privacy concerns of dealing with sensitive information, a common method that is often used in practice is the anonymization of personally identifiable data.
But, such approaches are susceptible to de-anonymization attacks~\cite{narayanan2008robust}.
That made researchers explore alternative methods.
To make the system immune to such attacks, privacy-preserving machine learning approaches have been developed~\cite{al2019privacy,li2016distributed}.
This particular privacy challenge is usually compounded by some auxiliary ones due to the presence of complex and often noisy data that, in practice, might typically consist of a combination of multiple types of data such as discrete, continuous, and categorical. 

\subsection{Synthetic Data Generation}
One of the most promising privacy-preserving approaches is \textit{Synthetic Data Generation~(SDG)}.
Synthetically generated data can be shared publicly without privacy concerns and provides many collaborative research opportunities, including for tasks such as building prediction models and finding patterns.
As SDG inherently involves a generative process, Generative Adversarial Networks~(GANs) \cite{goodfellow2014generative} attracted much attention in this research area, due to their recent success in other domains.

GANs are not reversible, i.e., one may not use a deterministic function to go from the generated samples to the real samples, as the generated samples are created using an implicit distribution of the real data. However, a naive usage of GANs for SDG does not guarantee the system being privacy-preserving by just relying on the fact that GANs are not reversible, as GANs are already proven to be vulnerable~\cite{hayes2019logan}. 

This problem becomes much more severe when such privacy violations can have serious consequences, such as when dealing with patient sensitive data in the medical domain. The primary ambiguity here is to understand how a system can claim to preserve the original data's privacy.
More precisely, how private is a system?
How much information is leaked during the training process?
Thus, \textit{there is a need to measure the privacy of a system} -- to be able to judge if a system is privacy-preserving or not.

\subsection{Differential Privacy}
\textit{Differential Privacy~(DP)}~\cite{dwork2014algorithmic} provides a mechanism to ensure and quantify the privacy of a system using a solid mathematical formulation.
Differential privacy recently became the de facto standard for statistical exploration of databases that contain sensitive private data.
The power of differential privacy is in its accurate mathematical representation, that ensures privacy, without restricting the model for accurate statistical reasoning. Furthermore, utilizing differential privacy, one can measure the privacy level of a system.

Differential privacy, as a strong notion of privacy, becomes crucial in machine learning and deep learning as the system by itself often employs sensitive information to augment the model performance for prediction.
Although many different attack types can jeopardize a machine learning model's privacy, it is the safest to assume the presence of a powerful adversary with complete knowledge of the pipeline, training, and model~\cite{shokri2015privacy}.
Hence protecting the system against such an adversary, or at the very least, measuring the upper bound of the privacy leakage in this scenario, gives us a full view of the system's privacy.
A fully differentially private system guarantees that the algorithm's training is not dependent on an individual's sensitive data.

One of the most adopted approaches to ensure differential privacy, under a modest \textit{privacy budget}, and model accuracy, is \textit{Differentially Private Stochastic Gradient Descent (DP-SGD)}, proposed in \cite{abadi2016deep}, which is the basis for many different research works \cite{xie2018differentially,acs2018differentially}.
At its core, DP-SGD (1) bounds the sensitivity of the algorithm on individuals by clipping the gradients, (2) adds Gaussian noise, and (3) performs the gradient descent optimization step. As of now, SGD under a DP regime is one of the most promising privacy-preserving approaches.  


Due to the rapid rise in the utilization of differential privacy, practical issues related to tracking the privacy budget became a subject of many discussions.
Despite its great advantages, the proposed notion of differential privacy has some disadvantages due to what is called the \textit{composition theorem}~\cite{dwork2014algorithmic,dwork2006calibrating}.
The composition theorem indicates that, for some composed mechanisms, the privacy cost simply adds up, leading to the discussion of privacy budget restriction.
This is problematic, especially in deep learning, as the model training is an iterative process, and each iteration adds to the privacy loss.
To tackle the shortcomings of the differential privacy definition, \textit{Rényi Differential Privacy~(RDP)} has been proposed as a natural relaxation
of differential privacy~\cite{mironov2017renyi}.
As opposed to differential privacy, RDP is a more robust notion of privacy that may lead to a more accurate and numerically stable computation of privacy loss. 

\subsection{Challenges with Medical Data}
There are \textit{four primary challenges in the synthetic data generation research in the medical domain}: (1) \textbf{Preserving the privacy.} The majority of the existing works do not train the model in a privacy-preserving manner. At best, they only try to address privacy with some statistical or machine learning-based measurements.
(2) \textbf{Handling discrete data}. When discrete data is involved, many of the methods using GANs face some difficulty since GAN models are inherently designed for generating continuous values. In particular, they especially fail when there is a mixture of continuous and discrete data.
(3) \textbf{Evaluation of synthetic data quality}. One of the challenging issues regarding the evaluation of GANs and the quality of synthetic data in a real-world domain is: How can we measure the quality of a synthetically generated data? However, this becomes particularly more critical in the medical domain since the use of low-quality synthetic data for building prediction models can have dire consequences and may even jeopardize human lives.
(4) \textbf{Temporal information and correlated features}. The temporal and local correlation between the features is often ignored. Incorporating such information is important in the medical domain since the disease incidents, or individual patients' records, often have meaningful temporal/correlation patterns. The quality of the generated data can be significantly improved by incorporating such dependencies in the data.

\subsection{Our Contributions}
To address the problems described before, here we propose a privacy-preserving framework that employs \textbf{R}ényi \textbf{D}ifferential \textbf{P}rivacy and \textbf{C}onvolutional \textbf{G}enerative \textbf{A}dversarial \textbf{N}etworks (RDP-CGAN). Our work makes the following contributions:

\begin{itemize}[leftmargin=*]
    \item We track and compute privacy loss with stable numerical computations and have \textit{tighter bounds on the privacy loss function, which leads to better performance under the sample privacy budget}. We achieve this by employing Rényi differential privacy in our model. 
    \item Our model can effectively \textit{handle discrete data} and capture the information from a \textit{mixture of discrete-continuous data} by creating a compact, unified representation through \textit{Convolutional Autoencoders~(CEs)} for unsupervised feature learning.  
    \item In addition to measuring the quality of synthetic data with statistical measurements, we also employ a  \textit{labeled synthetic data generation mechanism} for assessing the quality of the synthetic data in a supervised manner.
    \item To \textit{incorporate temporal and correlation dependencies in the data}, we utilize one-dimensional convolutional neural networks. 
    
\end{itemize}

\begin{table*}
  \caption{Comparison of methods based on different criteria.}
  \label{tab:methodscomparisoncriteria}
  \centering
  \resizebox{.9\textwidth}{!}{
\begin{tabular}{c||c|c|c|c|c|c}
\toprule
    
    Method & MedGAN & TableGAN & CorGAN & DPGAN & PATE-GAN & RDP-CGAN \\
     & \cite{choi2017generating} & \cite{park2018data}  & \cite{torfi2020corgan} & \cite{xie2018differentially} & \cite{jordon2018pate} & (Proposed) \\
    
    \midrule
    Privacy-Preserving & $\times$  &  $\times$  &   $\times$  & \checkmark  &   \checkmark   & \checkmark \\
    Does Not Require Labels (Annotations) & \checkmark  & $\times$     &   \checkmark  & \checkmark &   $\times$   & \checkmark\\
    Can Handle Mixed Data Types & \checkmark    & \checkmark  &    \checkmark   & $\times$ & $\times$  & \checkmark \\
    Capture Correlated and Temporal Information & $\times$   & \checkmark  &   \checkmark   & $\times$ & $\times$  & \checkmark \\
    \bottomrule
\end{tabular}
}
\end{table*}

We demonstrate that our model generates higher quality synthetic data under the same privacy budget. Likewise, it can provide a higher level of privacy under the same level of synthetic data quality.
We also empirically show that, regardless of privacy considerations, our proposed architecture generates higher quality synthetic data quality as it captures temporal and correlated information.

The rest of this part is organized as follows: Section~\ref{sec:related} investigate some prior research efforts related to synthetic medical data generation and differentiates our work from other existing works.
Section~\ref{sec:preleminary} describes preliminary concepts required to comprehend our proposed work.
Section~\ref{sec:framework} provides the structural and implementation details of the proposed system.
Section~\ref{sec:preleminary} demonstrates the comparison results with state-of-the-arts by privacy analysis and the quality of generated synthetic data.
Finally, Section~\ref{sec:conclusions} concludes our findings in this paper.

\section{Related Works}\label{sec:related}

There are a wide variety of works utilizing Differential Privacy for generating synthetic data.
A majority of them use the mechanism proposed in \cite{abadi2016deep} to train a neural network with differential privacy  based ipon gradient clipping for bounding the gradient norms and adding noise.
It follows the general mechanism proposed in \cite{dwork2014algorithmic}.
One of the most critical contributions in \cite{abadi2016deep} is introducing the privacy accountant that tracks the privacy loss.
Motivated by the success of this mechanism, in our approach, we extend our privacy-preserving framework using Rényi Differential Privacy~(RDP)~\cite{mironov2017renyi} as a new notion of DP to calculate the privacy loss.

The problem of synthetic data generation in the healthcare domain has seen a recent surge in the research efforts \cite{baowaly2018synthesizing,pollack2019creating,guan2018generation}.
Perhaps one of the earlier works in generating synthetic medical data is MedGAN~\cite{choi2017generating} in which there is no privacy-preserving mechanism being enforced.
It only uses GAN models to create synthetic data, and it is well known that GANs are prone to be attacked.
Hence, despite its good performance in generating data, it does not provide any kind of privacy guarantee.
One of the primary contributions of \textit{MedGAN} is to tackle the discrete data generation through denoising autoencoders~\cite{choi2017generating}.
One the other hand, utilization of synthetic patient population simulators such as Synthea~\cite{walonoski2017synthea} is minimal as it only relies on standard guidelines and does not model factors that may contribute to a predictive analysis \cite{chen2019validity}.
Another work is \textit{CorGAN}~\cite{torfi2020corgan} that tries to generate longitudinal event sequences by capturing feature correlations and temporal information. The \textit{TableGAN} approach~\cite{park2018data} also uses convolutional GANs to synthesize data by leveraging auxiliary classifier GANs.
Another work in this area is the \textit{CTGAN} model presented in \cite{xu2019modeling} that addresses tabular data which has a mixture of discrete and continuous features.
However, none of these approaches guarantees any sort of privacy during the data generation.
Hence, there is a good chance that many of these models can compromise the privacy of the original medical data, making these models vulnerable in real practice.
Accordingly, we will now discuss various privacy-preserving strategies.

One of the earliest works that addressed the privacy-preserving deep learning in the healthcare domain is \cite{beaulieu2019privacy}, that uses \textit{Auxiliary Classifier GANs}~\cite{odena2017conditional} to generate synthetic data.
It assumes having access to labeled data.
However, we aim to develop a model able to create synthetic data from both labeled and unlabeled real data.
\textit{PATE-GAN}~\cite{jordon2018pate} is one of the successful SDG approaches that use the \textit{Private Aggregation of Teacher Ensembles (PATE)}~\cite{papernot2016semi,papernot2018scalable} mechanism to ensure Differential Privacy.
Another framework related to this topic is \textit{DPGAN}~\cite{xie2018differentially}, which implements a mechanism similar to the one developed in \cite{abadi2016deep}, with the main difference of clipping weights instead of gradients.
We compare our work with PATE-GAN and DPGAN due to their success and relevance to the medical domain. One of the research effort associated with utilizing RDP is \cite{torkzadehmahani2019dp} in which the authors training a privacy-preserving model on MNIST dataset build on conditional GANs. The work proposed in \cite{tantipongpipat2019differentially} uses RDP to enforce differential privacy. However, they do not consider temporal information. Further, we claim that their system is not fully-differentially private as they only partially apply differential privacy. A general comparison for related methods is depicted in Table~\ref{tab:methodscomparisoncriteria}. The rows in Table~\ref{tab:methodscomparisoncriteria} describe important characteristics. First, we desire to investigate if a model \textit{preserves the privacy or not}. Second, it is crucial to assess the model's utility by checking if a model \textit{requires labels}. If a model does not need labels, it can be used for unsupervised synthetic data generation. Third, we target the approaches that can \textit{handle a mixture of data types}. At last, we are also interested in the models that can \textit{capture correlated and temporal information}.

To assess the privacy-preserving characteristics of our model, we compare our method to two of the prominent research efforts that guarantee differential privacy in the context of synthetic data generation, namely, DPGAN and PATE-GAN. PATE-GAN uses a modified version of a PATE mechanism. One issue with the original PATE is that its privacy cost depends on the amount data it requires for a labeling process incorporated in the mechanism. Noted that PATE mechanism uses some public dataset (similar to real private data) for its labeling mechanism.
Not only may this increase the privacy loss significantly, but the availability of such a labeled public dataset is a limiting assumption in a practical, real-world scenario.
Although PATE-GAN changed the PATE training paradigm so that it does not require public data, its basis is to train a student model by employing the generator output that may aggregate the error, and it still needs labeled data.
However, the advantage of PATE is that it provides a tighter privacy upper bound as opposed to the mechanism used in DPGAN.
Regarding the privacy, the advantage of our method compared to both these methods is that \textit{we use the RDP mechanism which provides a tighter privacy upper bound.}
Furthermore, our approach uses \textit{convolutional architectures which typically perform better than standard multilayer perception models} by capturing the correlated features and temporal information.
Furthermore, we use \textit{convolutional autoencoders that, in an unsupervised manner}, create a feature space that can effectively incorporate both discrete and continuous values.


\section{Preliminaries}\label{sec:preleminary}

\subsection{Autoencoders}

Autoencoder are a kind of neural network architectures that consists of two encode and decode functions.
$\boldsymbol{Enc(.)}:\mathbb{R}^n \rightarrow \mathbb{R}^d$ and $\boldsymbol{Dec(.)}:\mathbb{R}^d \rightarrow \mathbb{R}^n$ aim to transpose the input $\boldsymbol{x} \in \mathbb{R}^n$ to the latent space $\mathcal{I} \in \mathbb{R}^d$ and then reconstruct $\boldsymbol{\hat{x}} \in \mathbb{R}^n$.
In an ideal scenario, a perfect reconstruction of the original input data can be achieved, i.e., $x=\hat{x}$.
The autoencoders usually employ \textit{Binary Cross Entropy~(BCE)} and \textit{Mean Square Error~(MSE)} losses for binary and continuous inputs, respectively.

\begin{equation}\label{eq:bce}
BCE(\boldsymbol{x},\boldsymbol{\hat{x}})=-\frac{1}{m}\sum_{i}^{m}\left ( x_i log(\hat{x}_i) + (1- x_i) log(1  \hat{x}_i) \right )
\end{equation}

\begin{equation}\label{eq:mse}
MSE(\boldsymbol{x},\boldsymbol{\hat{x}})=\frac{1}{m}\sum_{i=1}^{m} \left |x_i - \hat{x}_i \right |^2
\end{equation}

We utilize autoencoders in our model for capturing low-dimensional representations for both discrete and continuous variables.

\subsection{Differential Privacy}

\textit{Differential Privacy} establishes a guarantee of individuals' privacy via measuring the privacy loss
(associated with any information release extracted from a database) by a mathematical definition~\cite{dwork2006calibrating,dwork2014algorithmic}.
The most widely adopted definition is the $(\epsilon,\delta)$-differential privacy.

\begin{definition}[$(\epsilon,\delta)$-DP]\label{def:dp}
The randomized algorithm $\mathcal{A}: \mathcal{X} \rightarrow \mathcal{Q}$, is $(\epsilon,\delta)$-differentially private for all query outcomes $\mathcal{Q}$ and all neighbor datesets $D$ and ${D}'$ if:

$$Pr[\mathcal{A}(D) \in \mathcal{Q}] \leq e^{\epsilon} Pr[\mathcal{A}({D}') \in \mathcal{Q}] + \delta$$

\end{definition}
Two datasets $D$, ${D}'$ that only differ by one record
(e.g., a patients' record) are called \textit{neighbor datasets}.
The notion of neighboring datasets emphasizes the sensitivity of any individual private data.
The parameters $(\epsilon,\delta)$ denote the privacy budget, i.e., being differentially private does not mean we have absolute privacy.
It only indicates our confidence level of privacy, given the $(\epsilon,\delta)$ parameters.
The smaller the $(\epsilon,\delta)$ parameters are, the more confident we become about our algorithm's privacy as $(\epsilon,\delta)$ indicates the privacy loss by definition. 

The $(\epsilon,\delta)$-DP with $\delta=0$ is called $\epsilon$-DP which is the initially proposed definition \cite{dwork2006calibrating}.
It provides a more substantial promise of privacy as even a tiny amount of $\delta$ might be a dangerous privacy violation due to the distribution shift~\cite{dwork2006calibrating}.
One of the most important reasons in the usage of $(\epsilon,\delta)$-DP is the applicability of advanced composition theorems~\cite{dwork2014algorithmic,dwork2010boosting}.

\begin{theorem}[Strong Composition~\cite{dwork2014algorithmic}]\label{def:strongcomposition}
Assume we execute a $k$-fold adaptive composition of a mechanism with $(\epsilon,\delta)$-DP. Then, the composite mechanism is $({\epsilon}',k{\delta}'+\delta)$-DP for ${\delta}'$, in which ${\epsilon}'=\sqrt{2kln(1/{\delta}')}\epsilon+k\epsilon(e^{\epsilon}-1)$~\cite{dwork2014algorithmic}. $\blacktriangle$

\end{theorem}

\subsection{Rényi Differential Privacy}

The strong composition~(see Definition~\ref{def:strongcomposition}) enables tighter upper bound for compositions of $(\epsilon,\delta)$-DP steps compared to the basic composition theorem~\cite{dwork2014algorithmic}.
One particular problem with the advanced composition theorem is that iterating this process leads to the rapid growth of parameters as each employment of the advanced composition theorem
(each step of the iterative process) lead to a various selection of possible $(\epsilon(\delta),\delta)$ values.
To address some of the shortcomings of the $(\epsilon,\delta)$-DP definition, the \textit{Rényi Differential Privacy~(RDP)} has been proposed in \cite{mironov2017renyi} based on Rényi divergence (Eq.~\ref{def:rd}).

\begin{definition}[Rényi divergence of order $\alpha$ \cite{renyi1961measures}]\label{def:rd}

The Rényi divergence of order $\alpha$ of a distribution $P$ from the distribution $P'$ is defined as:

$$D_\alpha(P||P')=\frac{1}{\alpha - 1} log \left (E_{P'(x)}\left ( \frac{P(x)}{P'(x)} \right )^{\alpha}\right )$$

\end{definition}

The Rényi divergence, as a generalization of the Kullback–Leibler divergence, is precisely equal to the Kullback–Leibler divergence for the order $\alpha=1$. The special case of $\alpha=\infty$ equals to:

$$D_\infty(P||P')= log \left ( sup_{P'(x)} \frac{P(x)}{P'(x)}\right )$$

which is the log of the maximum ratio of the probabilities over $P'(x)$. The order of $\alpha=\infty$ establishes the connection between the Rényi divergence and the $\epsilon$-DP. A randomized mechanism $\mathcal{A}$ is $\epsilon$-differentially private if for two neighbor datasets $D$ and ${D}'$, we have:

$$D_\infty(\mathcal{A}(D)||\mathcal{A}({D}')) \leq \epsilon$$

Now, based on these definitions, let us define Rényi
differential privacy (RDP) \cite{mironov2017renyi} as a new notion of differential privacy.

\begin{definition}[Rényi Differential Privacy~(RDP) \cite{mironov2017renyi}]\label{def:rdp}

The randomized algorithm $\mathcal{A}: \mathcal{D} \rightarrow \mathcal{U}$ is $(\alpha,\epsilon)$-RDP if, for all neighbor datasets $D$ and ${D}'$, we have:

$$D_\alpha(\mathcal{A}(D)||\mathcal{A}({D}')) \leq \epsilon$$

\end{definition}

There are two essential properties of the RDP definition (Definition~\ref{def:rdp}) that are required here.

 \begin{proposition}[Composition of RDP \cite{mironov2017renyi}]\label{def:rdp-p1}
 
 If $\mathcal{A}: \mathcal{X} \rightarrow \mathcal{U}_1$ is $(\alpha,\epsilon_{1})$-RDP and $\mathcal{B}: \mathcal{U}_1 \times \mathcal{X} \rightarrow \mathcal{U}_2$ is $(\alpha,\epsilon_{2})$-RDP, then the mechanism $(\mathcal{M}_1,\mathcal{M}_2) $, where $\mathcal{M}_1 \sim \mathcal{A}(\mathcal{X})$ and $\mathcal{M}_2 \sim \mathcal{B}(\mathcal{M}_1,\mathcal{X})$, satisfies the $(\alpha,\epsilon_{1}+\epsilon_{2})$-RDP conditions.$\blacktriangle$

\end{proposition}

 \begin{proposition}\label{def:rdp-p2}
 If the mechanism $\mathcal{A}: \mathcal{X} \rightarrow \mathcal{U}$ is $(\alpha,\epsilon)$-RDP then it also obeys $(\epsilon + \frac{log(1/\delta)}{(\alpha-1)},\delta)$-DP for any $0 < \delta < 1$ \cite{mironov2017renyi}. $\blacktriangle$

\end{proposition}

The above two propositions form the basis for preserving privacy in our approach. Proposition~\ref{def:rdp-p1} determines the calculation of the privacy cost in our framework as a composition of two structures~(autoencoder and convolutional GAN) and Proposition~\ref{def:rdp-p2} is applicable when one wants to calculate the extent to which our system is differentially private based on the traditional $(\epsilon, \delta)$ notion~(see Definition~\ref{def:dp}).

\section{Privacy-Preserving Framework}\label{sec:framework}

\begin{figure*}[ht]
\centering
\includegraphics[width=0.85\textwidth]{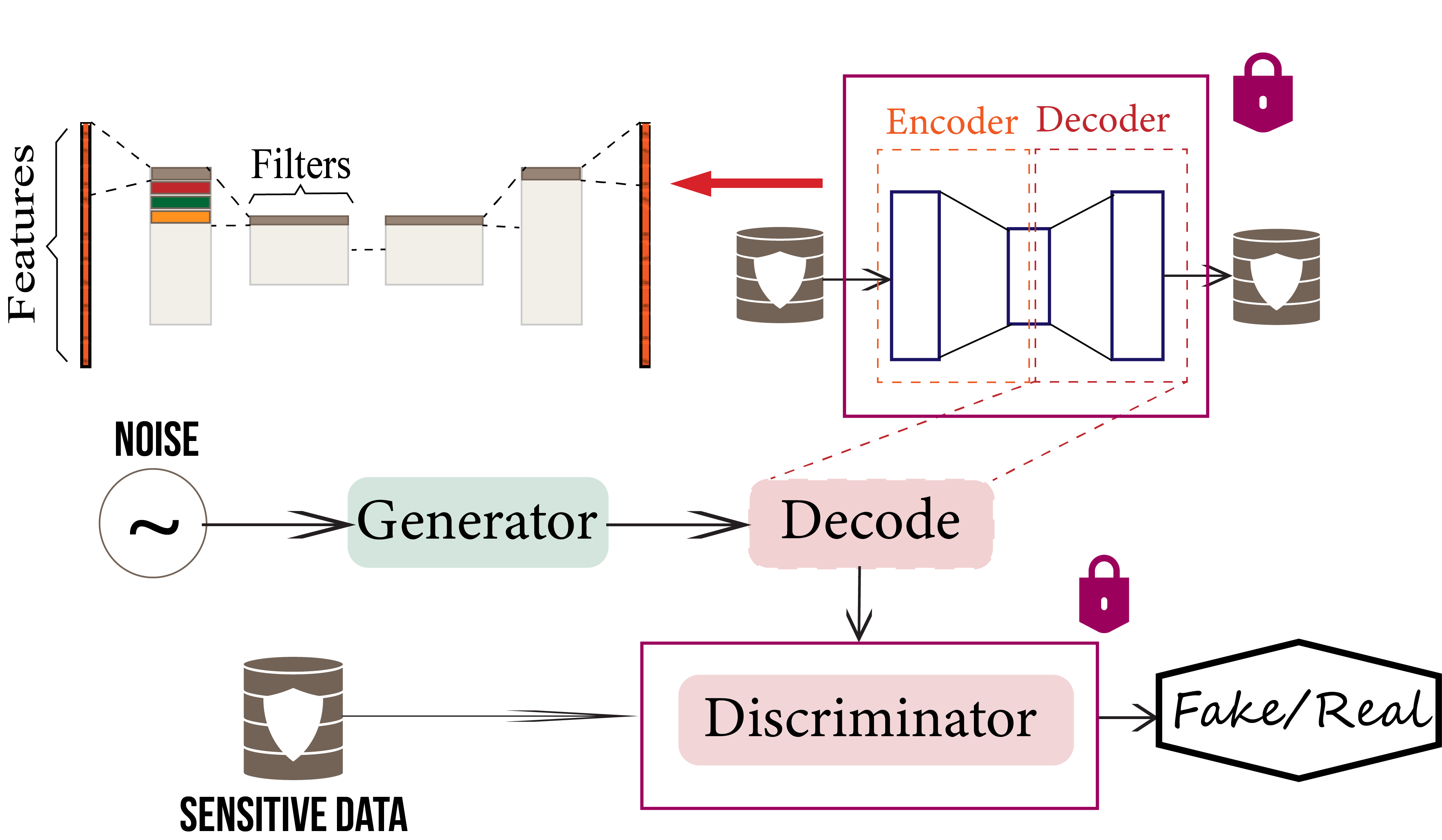}
\caption{The proposed RDP-CGAN framework. In the Autoencoder, the encoders and decoders are formed with convolutional layers. Discriminator and Generator have architectures similar to that of the encoder and decoder, respectively.}
\label{fig:rdp-model}
\end{figure*}

In our work, we build a privacy-preserving GAN model using Rényi differential privacy.
However, since it is well known that the GAN models typically have a poor performance in generating non-continuous data \cite{hjelm2017boundary}, we utilize autoencoders~\cite{kingma2013auto} to aid the GAN model by creating a continuous feature space to represent the input.
At the same time, the GAN aims to generate high-fidelity synthetic data in a privacy-preserving manner. 
The autoencoder will transform the input space into a continuous space, whether we have a discrete, continuous, or a mixture of both types as the input.
Thus, the autoencoder acts as a bridge between the GAN model and the non-continuous data present in the medical domain.

Furthermore, the majority of existing research efforts in generating synthetic data~\cite{choi2017generating} ignore the features' local correlations or temporal information by using multilayer perceptrons, which are inconsistent with real-world scenarios such as disease progression.
To remedy this drawback, for both autoencoder and GAN architecture~(generator \& discriminator), we use convolutional neural networks.
In our approach, we employ \textit{One Dimensional Convolutional Neural Networks~(1D-CNNs)} to capture correlated input features' patterns as well as temporal information and incorporate them within an autoencoder based framework.

The proposed framework is depicted in Fig.~\ref{fig:rdp-model}.
The inputs to the \textit{generator} $\boldsymbol{G}:\mathbb{R}^r \rightarrow \mathcal{D}_{g}^d$ and the \textit{discriminator} $\boldsymbol{D}:\mathbb{R}^n \rightarrow \mathcal{D}_{d}$ are the random noise $\boldsymbol{z} \in \mathbb{R}^r$ sampled from $\mathcal{N}(0,1)$ and the real data $\boldsymbol{x} \in \mathbb{R}^n$.
$\mathcal{D}_{g}^d$ and $\mathcal{D}_{d}$ are the generator and discriminator domains which are usually $\{0,1\}$ and $[-1,1]^d$, respectively.
In RDP-CGAN, as opposed to the regular GAN's training procedure, the generated fake data is \textit{decoded} before being fed to the discriminator.
The decoding is done by feeding the fake data to a pre-trained autoencoder.

\begin{algorithm} 
    \caption{Private Convolutional Autoencoder Pre-Training}
    \label{alg:rdp-sdg-ae}
    \begin{algorithmic}[1] 
        \State \textbf{Inputs \& Parameters}:
Real data $X=\{x_i\}_{i=1}^N$, learning rate $\eta$, network weights $\theta$, number of epochs for autoencoder training $n_{ae}$, the norm bound $C$, the additive noise standard deviation $\sigma_{ae}$.

\For{\(i=1\ldots n_{ae}\)}
        \State Sample a mini-batch of $n$ examples. $\mathcal{X}=\{x_i\}_{i=1}^n$.
        \State Partition $\mathcal{X}$ into $\mathcal{X}_1,\ldots,\mathcal{X}_r$ where $r=\left \lfloor \frac{n}{k} \right \rfloor$.
        \For{\(l=1\ldots r\)}
        \State $\mathcal{L}= BCE(\mathcal{X}_l,\hat{\mathcal{X}_l})$, $\hat{\mathcal{X}_l} = Dec(Enc(\mathcal{X}_l))$
        \State $\boldsymbol{g_{\theta,l}} \leftarrow \bigtriangledown_{\theta} \mathcal{L}(\theta,\mathcal{X}_l)$
\State \(\hat{\boldsymbol{g_{\theta,l}}} \leftarrow \boldsymbol{g_{\theta,l}} / max\left ( 1, \frac{\left \| \boldsymbol{g_{\theta,l}}() \right \|_2}{C} \right )\)
\EndFor
\State \(\hat{\boldsymbol{g_\theta}} \leftarrow \frac{1}{r}\sum_{l=1}^r \left (\boldsymbol{\hat{g_{\theta,l}}} + \mathcal{N}(0,\sigma_{ae}^2C^2 \mathbb{I}) \right )\)
        \State Update: $\hat{\theta} \leftarrow \theta - \eta \hat{\boldsymbol{g_\theta}}$
        
\EndFor
    
    \end{algorithmic}
\end{algorithm}

\begin{algorithm} 
    \caption{Private 1-D Convolutional GAN Training}
    \label{alg:rdp-sdg-cgan}
    \begin{algorithmic}[1] 
        \State \textbf{Inputs \& Parameters}:
Real data $X=\{x_i\}_{i=1}^N$, random noise $\boldsymbol{z}$ where $z_i \sim \mathcal{N}(0,1)$., learning rate $\eta$, discriminator weights $\omega$, generator weights $\psi$, number of epochs for GAN training $n_{gan}$, the norm bound $C$, the additive noise standard deviation $\sigma_{gan}$, the number of discriminator training steps per one step of generator training $n_d$.

    
    \For{\(j=1\ldots n_{gan}\)}
        \For{\(k=1\ldots n_d\)}
        \State Take a mini-batch from private data $\mathcal{X}=\{x_i\}_{i=1}^n$
        \State Sample a mini-batch $\mathcal{Z}=\{z_i\}_{i=1}^n$
        \State Partition real data mini-batches into $\mathcal{X}_1,\ldots,\mathcal{X}_r$
        \State Partition noise data mini-batches into $\mathcal{Z}_1,\ldots,\mathcal{Z}_r$
        \For{\(l=1\ldots r\)}
        \State $x_i \in \mathcal{X}_l$ and $ z_i \in \mathcal{Z}_l$
        \State $\mathcal{L}$= $\frac{1}{k}\sum_{i=1}^{k}\left ( D(x_{i}) - D(\boldsymbol{Dec}(G(z_{i}))) \right )$
        \State $\boldsymbol{g_{\omega,l}} \leftarrow \bigtriangledown_{\omega} \mathcal{L}(\omega,\mathcal{X}_l)$
\State \(\hat{\boldsymbol{g_{\omega,l}}} \leftarrow \boldsymbol{g_{\omega,l}} / max\left ( 1, \frac{\left \| \boldsymbol{g_{\omega,l}}() \right \|_2}{C} \right )\)
\EndFor
\State \(\hat{\boldsymbol{g_{\omega}}} \leftarrow \frac{1}{r}\sum_{l=1}^r \left (\boldsymbol{\hat{g}_{\omega,l}} + \mathcal{N}(0,\sigma_{gan}^2C^2 \mathbb{I}) \right )\)
        \State Update: $\hat{\omega} \leftarrow \omega - \eta \hat{\boldsymbol{g_{\omega}}}$
        \EndFor
    
        \State Sample $\{z_i\}_{i=1}^n$ from noise prior
        \State $\mathcal{L}$= $-\frac{1}{n}\sum_{i}^{n}\left ( D(Dec(G(z_i))) \right )$
        \State $\boldsymbol{g_{\psi}} \leftarrow \bigtriangledown_{\psi} \mathcal{L}(\psi,\mathcal{Z})$
        \State Update: $\hat{\psi} \leftarrow \psi - \eta \boldsymbol{g_{\psi}}$
    
    \EndFor
    
    \end{algorithmic}
\end{algorithm}

\subsection{Convolutional Autoencoder}

We build a \textit{one-dimensional convolutional autoencoder~(1D-CAE)} based architecture.
The main goal of such a 1D-CAE is to:
\textbf{(1)} capture the correlation between neighboring features,
\textbf{(2)} represent a new compact feature space,
\textbf{(3)} transform the possible discrete records to a new continuous space, and
\textbf{(4)} simultaneously model discrete and continuous phenomena, which is a challenging problem~\cite{xu2019modeling}. 

We enforce privacy by adding noise, and by clipping gradients of both the encoder and decoder
(see Algorithm~\ref{alg:rdp-sdg-ae}, \textit{Autoencoder Pretraining Step}), as in some previous research works~\cite{acs2018differentially,abay2018privacy}.
One might claim adding noise to the decoder part should suffice as the discriminator only has access to the decoder part of the CAE.
We assert that such an action might jeopardize the privacy of the model since we cannot rely on the autoencoder training to be trusted and secure as it has access to the real data.
The details of autoencoder pretraining, demonstrated in Algorithm~\ref{alg:rdp-sdg-ae} are as follows:

\begin{itemize}[leftmargin=*]
    \item We pre-train the autoencoder for $n_{ae}$ steps (which will be determined based on the privacy budget $\epsilon$), and hence it varies by changing the desired level of $\epsilon$.
    \item We divide a mini-batch into several micro-batches.
    \item For each micro-batch of size 1, we calculate the loss (\textit{line 7}), calculate the gradients (\textit{line 8}), and clip the gradients to bind the model's sensitivity to the individuals (\textit{line 9}).
    The operation $\left \| \boldsymbol{g_i}() \right \|_2$ indicates the $\ell_{2}$ norm over the micro-batch gradients, and $C$ is an arbitrary upper bound for the gradients' norm.
    \item The Gaussian noise ($\mathcal{N}(0,\sigma_{ae}^2C^2 \mathbb{I})$)will then be independently added to the gradients of each micro-batch and will be aggregated (\textit{line 10}). 
    \item The last step is where the optimizer performs the parameter update (\textit{line 11}).
\end{itemize}

\subsection{Convolutional GAN}

The pseudocode for the Convolutional GAN is given in Algorithm~\ref{alg:rdp-sdg-cgan}, under the procedure \textit{GAN Training Step}.
As can be observed in Algorithm~\ref{alg:rdp-sdg-cgan}, we only enforce differential privacy on the discriminator as only it has access to the real data.
To avoid the issue of mode collapse~\cite{gulrajani2017improved}, we train the CGAN model, using Wasserstein GAN~\cite{arjovsky2017wasserstein} since it is an efficient approximation of \textit{Earth Mover~(EM)} distance and is shown to be robust to the mode collapse problem during model training~\cite{arjovsky2017wasserstein}.

The \emph{Earth-Mover (EM)} distance or Wasserstein-1 distance represents the minimum price in transforming the generated data distribution $\boldsymbol{P_g}$ to the real data distribution $\boldsymbol{P_r}$:

  \begin{equation}
    W(\mathbb{P}_x, \mathbb{P}_g) = \inf_{\vartheta \in \Pi(\mathbb{P}_x ,\mathbb{P}_g)} E_{(x, y) \sim \vartheta}[\:\|x - y\|\:]~,
    \label{eq::WGAN}
  \end{equation}
  $\vartheta(x,y)$ refers to how much ``mass'' should
  be moved from $x$ to $y$ to transform $\mathbb{P}_x$ to $\mathbb{P}_g$.

However, as the infimum in
\eqref{eq::WGAN} is intractable, based on the the Kantorovich-Rubinstein duality \cite{villani2008optimal}, the WGAN propose the Eq.~\ref{eq:KR} optimization:

\begin{equation} \label{eq:KR}
W(\mathbb{P}_x, \mathbb{P}_g) = \sup_{\|f\|_L \leq 1} E_{x \sim \mathbb{P}_x}
[f(x)] - E_{x \sim \mathbb{P}_g}[f(x)]
\end{equation}
For simplicity of the definition, infimum and supremum indicate the greatest lower bound and the least upper bound, respectively.

 \begin{definition}[1-Lipschitz functions]\label{def:Lipschitz} Given two metric spaces $(X, d_X)$ and $(Y, d_Y)$, where $d$ denotes the metric (e.g., distance metric), the function $f: \mathbb{X} \rightarrow \mathbb{Y}$ is called K-Lipschitz if:

\begin{equation} \label{eq:K-Lipschitz}
\forall (x,x') \in \mathbb{X}, \exists K \in \mathbb{R}: d_Y(f(x),f(x'))\leq K d_X(x,x')
\end{equation}

\end{definition}

Using the distance metric and $K=1$, Eq.~(\ref{eq:K-Lipschitz}) is equivalent to:

\begin{equation} \label{eq:1-Lipschitz}
\forall (x,x') \in \mathbb{X}: |f(x) - f(x')|\leq |x - x'|
\end{equation}

It is clear that for \textit{computing the Wasserstein distance, we should find a 1-Lipschitz function~(see Definition~\ref{def:Lipschitz} and Eq.~\ref{eq:1-Lipschitz}.}.
The approach is to build a neural model to learn it.
The procedure is to construct a discriminator D without the Sigmoid function, and output a scalar instead of the probability of confidence.

Regarding the privacy considerations, the generator has no access to real data directly (although it has access to the gradients from the discriminator); only the discriminator will have access to it.
We propose only to train the discriminator under differential privacy for this aim.
More intuitively, this approach considers the post-processing theorem in differential privacy as follows.

\begin{theorem}[Post-processing~\cite{dwork2014algorithmic}]
Assume $\boldsymbol{D}:\mathbb{N}^{|\mathcal{X}|} \rightarrow \mathbb{D}$ is an algorithm which is $(\epsilon, \delta)$-differentially private and $\boldsymbol{G}:\mathbb{D} \rightarrow \mathbb{O}$ is some arbitrary function operates on .
Then $\boldsymbol{G} \circ \boldsymbol{D}:\mathbb{N}^{|\mathcal{X}|} \rightarrow \mathbb{O}$ is $(\epsilon, \delta)$-differentially private as well~\cite{dwork2014algorithmic}.$\blacktriangle$
\end{theorem}

Following the argument above, we consider $\boldsymbol{D}$ and $\boldsymbol{G}$ as discriminator and generator functions, respectively.
Since the generator is an arbitrarily randomized mapping on top of the discriminator, enforcing the differential privacy on the discriminator suffices to guarantee that the overall system ensures differential privacy, and we do not need to train a private generator as well.
The generator training procedure is given in Algorithm~\ref{alg:rdp-sdg-cgan}, \textit{lines 17-20}.
As can be observed, we do not have a private generator and the loss function is the regular generator loss function of the WGAN method~\cite{arjovsky2017wasserstein}.

\subsection{Architecture Details}

For the GAN  discriminator, we used five convolutional layers similar to the autoencoder (encoder), except for the last layer that we have another dense layer, of output size 1, for decision making.
For all layers, we used PReLU activation~\cite{he2015delving}, except for the last layer that does not use any activation.
For the generator, we used \textit{transposed convolutions} (also called \textit{fractionally-strided convolutions}) similar to the ones used in \cite{radford2015unsupervised}.
However, we use 1-D transposed convolutions. The GAN generator has an input noise of size \textit{100}, and its output size is set to 128. 


In the autoencoder and for the encoder, we used an architecture similar to the GAN discriminator, except we discard the last layer of the discriminator.
For the decoder, as done in the generator, we used \textit{transposed convolutional layers} in the reverse order of the ones used for the encoder.
For the encoder, we used \textit{PReLU} activation layers except for the last layer of the encoder where \textit{Tanh} was used to match the generator output.
For the decoder, we also used \textit{PReLU} activation except for the last layer where a \textit{Sigmoid} activation function was used to bound the range of output data to the $[0,1]$ to ideally reconstruct the discrete range of $\{0,1\}$ aligned with the input data.
The decoder input size (the encoder output size) is equal to the GAN's generator output dimension.

Since our model inputs sizes may change for different datasets, we modify the input pipeline of our architecture by varying the dimensions of convolution kernels, stride for the GAN discriminator, and the autoencoder, to match the new dimensionality.
The GAN generator \textit{does not require any change} since, for all experiments, we used the same noise dimension as mentioned above.

\subsection{Privacy Loss}

As tracking the RDP privacy accountant~\cite{abadi2016deep} is computationally more precise than the regular DP, \textit{we based our privacy loss calculation on RDP}.~Then, the RDP computations can be transformed to DP based on the proposition~\ref{def:rdp-p2}. The Gaussian noise additive procedure is also called \textit{Sampled Gaussian Mechanism~(SGM)}~\cite{mironov2019r}.~For tracking privacy loss, we use the following Theorem.

\begin{theorem}[SGM privacy loss~\cite{mironov2019r}] Assume $D$ and $D'$ are two neighbor datasets and $\mathcal{G}$ is a SGM applied to a function $f$ with $\ell_{2}$-sensitivity of one.~If we set the following:

$$\mathcal{G}(D) \sim \varphi_{1} \overset{\Delta}{=} \mathcal{N}(0,\sigma^2)$$
$$\mathcal{G}(D') \sim \varphi_{2} \overset{\Delta}{=} (1-q)\mathcal{N}(0,\sigma^2) + q \mathcal{N}(1,\sigma^2)$$

where $\varphi_{1}$ and $\varphi_{2}$ are PDFs, then, the mechanism $\mathcal{G}$ satisfies $(\alpha,\epsilon)$-RDP for:

$$\epsilon \leq \frac{1}{\alpha - 1} log(max\left \{ A_\alpha,B_\alpha \right \})$$

where $ A_{\alpha} \overset{\Delta}{=} \mathbb{E}_{x \sim \varphi_{1}}\left [ \left ( \varphi_{2} / \varphi_{1} \right )^\alpha \right ]$ and $ B_{\alpha} \overset{\Delta}{=} \mathbb{E}_{x \sim \varphi_{2}}\left [ \left ( \varphi_{1} / \varphi_{2} \right )^\alpha \right ]$. $\blacktriangle$

\end{theorem}

It can be proven that $A_{\alpha} \leq B_{\alpha}$ and the RDP privacy computation can solely be focused on upper bounding $A_{\alpha}$ which can be calculated with a closed-form bound and numerical calculations~\cite{mironov2019r}.~We use the same numerical calculations here.~However, that bounds $\epsilon$ for each step.~The overall bound of $\epsilon$ for the entire training process can be calculated by $\epsilon_{total} = num\_steps \times \epsilon$ for any given $\alpha$~(see proposition~\ref{def:rdp-p1}).~To determine the tighter upper bound, we try multiple $\alpha$ values and use the minimum $(\epsilon)$ and its associated $\alpha$~(obtained by RDP privacy accountant) to compute the $(\epsilon,\delta)$-DP employing proposition~\ref{def:rdp-p2}.

\textbf{System privacy budget:} One important question is: how do we calculate the $(\alpha,\epsilon)$-RDP for the \textit{combination of an autoencoder and GAN training?} Consider proposition~\ref{def:rdp-p1}, in which $\mathcal{X}$, $\mathcal{U}_1$, and $\mathcal{U}_2$ are the input and output spaces of the autoencoder and the output space of the discriminator. The mechanisms $\mathcal{A}$ and $\mathcal{B}$ are the autoencoder and discriminator, respectively.~Consider $\mathcal{U}_1$ is what the discriminator observes after decoding of the fake samples, and $\mathcal{X}$ is the space of real inputs. Hence, the proposition~\ref{def:rdp-p1} directly results in having the whole system with $(\alpha,\epsilon_{ae}+\epsilon_{gan})$-RDP by fixing the $\alpha$.~But we cannot guarantee that we can have a fixed $\alpha$, and the RDP has a budget curve parameterized by it.~The procedure of fixing $\alpha$ is as follows.

\begin{itemize}[leftmargin=*]
    \item Assume we have two systems $S_1$ and $S_2$ such that one is the autoencoder, and the other is the GAN.~Hence we have two systems which $(\alpha_1,\epsilon_1)$-RDP and $(\alpha_2,\epsilon_2)$-RDP.
    \item Without loss of generality, we assume $\epsilon_1 \leq \epsilon_2$.~We pick $\alpha_{total}=\alpha_2$.~Now we have system $S_2$ which $(\alpha_{total},\epsilon_2)$-RDP.
    \item For $S_1$ we pick $\alpha_{total}=\alpha_1$ and calculate $\epsilon'$ that $\epsilon \leq \epsilon'$.
    \item Now, the total system $(S_1,S_2)$ satisfies $(\alpha_{total},\epsilon_2+\epsilon')$-RDP.
\end{itemize}


\section{Experiments}

In this section, we will first present the details of the experimental setup and then report the results obtained in various experiments and compare with different methods available in the literature.

\subsection{Experimental Setup}
Here, we split the dataset to train $\mathcal{D}_{tr}$ and test sets $\mathcal{D}_{te}$. We utilize $\mathcal{D}_{tr}$ to train the models, and then generate symthesized samples $\mathcal{D}_{syn}$ using the trained model.
We set $|\mathcal{D}_{syn}|=|\mathcal{D}_{tr}|$.
Recall from Section \ref{sec:framework} that although for different datasets we alter the architectures associated with the input size, the encoder output space (decoder input space) and the generator output space will always have the same dimensionality. Furthermore, the encoder input space, the decoder output space, and the discriminator input space will also have the same dimensionality.
It is worth noting that all of the reported $\epsilon$ values are associated with the $(\epsilon,\delta)$-DP definition with $\delta=10^{-5}$ unless otherwise stated.

Both the convolutional autoencoder and convolutional GAN were trained with the \textit{Adam optimizer}~\cite{kingma2014adam} with the learning rate of $0.005$, with a mini-batch size of \textbf{64}.
For both generator and discriminator we used Batch Normalization (BN)~\cite{ioffe2015batch} to improve the training.
We used one GeForce RTX 2080 NVIDIA GPU for our experiments.

We compare our framework with various different methods. Depends on the experiments, some of those methods may or may not be used for comparison depending on their characteristics. For further details, please see Table~\ref{tab:methodscomparisoncriteria}.

    

\subsection{Datasets}\label{sub:datasets}
To evaluate this performance, we used the following datasets for our experiments:

 \begin{enumerate} [leftmargin=*]
\item \textbf{MIMIC-III~\cite{johnson2016mimic}:} This dataset consists of the medical records from around 46K patients. The \textit{MIMIC-III} dataset~\cite{johnson2016mimic} includes the medical records of 46K patients.

\item \textbf{Kaggle Cervical Cancer~\cite{fernandes2017transfer}:} This dataset covers patients records and aims to classify the cervical cancer. There are multiple attributes in the dataset of discrete, and continuous types and one attribute associated with the class label~(e.g.,~Biopsy as the target for classification).

\item \textbf{UCI Epileptic Seizure Recognition~\cite{andrzejak2001indications}:} In the UCI Epileptic Seizure Recognition dataset~\cite{andrzejak2001indications}, the task is to classify the seizure activity. The features are the values of the Electroencephalogram~(EEG) records at various time points. 

\item \textbf{PTB Diagnostic ECG~\cite{bousseljot1995nutzung}:} The electrocardiograms~(ECGs) in this dataset~\cite{bousseljot1995nutzung} are used to classify the heart activity as normal or abnormal. This dataset contains 14552 samples, out of which 4046 are classified as normal, and 10506 are abnormal activities. 

\item \textbf{Kaggle Cardiovascular Disease~\cite{kagglecardio}:} This Kaggle datasetis used to determine if a patient has cardiovascular disease or not from a variety of features such as age, systolic blood pressure and diastolic blood pressure. This dataset has a mixture of discrete and continuous feature types as well.

\item \textbf{MIT-BIH Arrhythmia~\cite{moody2001impact}:} The MIT-BIH Arrhythmia database~\cite{goldberger2000physiobank} includes ambulatory ECG recordings, created from 47 subjects. records obtained from a mixed group of inpatients (requires overnight hospitalization) and outpatients (usually do not require overnight hospitalization) at Boston's Beth Israel Hospital. Unlike other datasets in this set of experiments, for the MIT-BIH Arrhythmia database, we are dealing with a multicategory classification task. The ECG signals correspond to normal~(negative) and abnormal cases~(positive). There are a total of five classes in this dataset. Class zero is associated with normal cases. The rest of the classes are associated with different arrhythmias and myocardial infarction which we call abnormal.

\end{enumerate}

\subsection{Baseline Comparison Methods}

We compare our model with different benchmark methods~(see Table~\ref{tab:methodscomparisoncriteria}). Depends on the nature of experiments, the models are used as benchmarks. For example, if the experiments are associated with different privacy settings, the models that do not preserve the privacy, will not be used for comparison.

 \begin{enumerate} [leftmargin=*]
\item \textbf{MedGAN~\cite{choi2017generating}:} MedGAN consists of an architecture that is designed to generate discrete records. It has an autoencoder and a vanilla GAN. MedGAN does not preserve privacy. MedGAN can be used for unsupervised synthetic data generation. We used MedGAN open-source implementation\footnote{\url{https://github.com/mp2893/medgan}}.

\item \textbf{TableGAN~\cite{park2018data}:} TableGAN consists
of three components: A generator, a discriminator, and an auxiliary classifier that aims to augment the semantic integrity of synthesized data. Similar to our method, TableGAN uses Convolutional Neural Networks. TableGAN requires labeled training data and will be used in experiments with a supervised setting. We used TableGAN open-source implementation\footnote{\url{https://github.com/mahmoodm2/tableGAN}}.

\item \textbf{DPGAN~\cite{xie2018differentially}:} DPGAN enforces differential privacy by clipping the weights rather than the gradients. It uses WGAN to stabilize the training and use multi-layer perceptions for its architecture. DPGAN does not require labeled data and hence will be used in both supervised and unsupervised settings. We used DPGAN open-source implementation~\footnote{\url{https://github.com/illidanlab/dpgan}}.

\item \textbf{PATE-GAN~\cite{jordon2018pate}:} PATE-GAN proposes a differentially private model formed based on the modification of the PATE mechanism. We utilized our own implementation of the PATE-GAN as its code was not available publicly.

\end{enumerate}

\subsection{Unsupervised Synthetic Data Generation}

We chose electronic health records after converting them to a high-dimensional binary discrete dataset to demonstrate the privacy-preserving potential of the proposed model. Here, we have $\mathcal{D}_{tr} \in \{0,1\}^{R \times |\mathcal{T}|}$, $\mathcal{D}_{te} \in \{0,1\}^{T \times |\mathcal{T}|}$, and $\mathcal{D}_{syn} \in \{0,1\}^{S \times |\mathcal{T}|}$, where $|\mathcal{T}|$ represents the feature size.~The goal is to capture the information from a real private dataset, and use that to synthesize a dataset with similar distribution in a privacy-sensitive manner. In this section, we can compare our method with models that do not require labeled data (for unsupervised synthetic data generation) and are also privacy-reserving method (see Table~\ref{tab:methodscomparisoncriteria}). In the experiments of this section, to pretrain the autoencoder, we used Eq.~\ref{eq:bce} as the loss function.

\textbf{Dataset Construction:} We used MIMIC-III dataset~\cite{johnson2016mimic} as an example for \textit{high-dimensional} and \textit{discrete} dataset. From MIMIC-III,  We extracted and used 1071 unique ICD-9 codes.~The dataset has different discrete variables (e.g., diagnosis codes, procedure, etc.).~Assuming there are $\boldsymbol{|V|}$ discrete variables, we can represent the patient's record using a binary vector $\boldsymbol{v} \in {\{0,1\} }^{|V|}$. The $i^{th}$ variable is set to one if it is available in the patient's record.~In our experiments, $\boldsymbol{|V|}=1071$.

    

\begin{figure}[ht]
\centering
\includegraphics[width=0.49\textwidth]{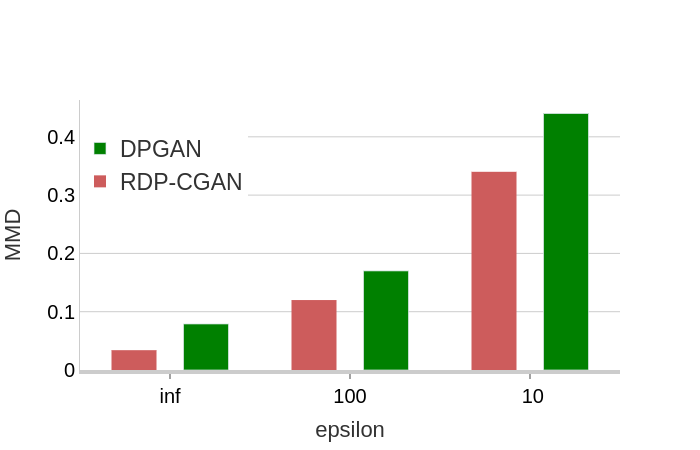}
\caption{The comparison of generated and real data distributions. Lower MMD score indicates a higher distribution similarity and hence, a better model.}
\label{fig:mimicmmd}
\end{figure}

\begin{figure}[ht]
\centering
\includegraphics[width=0.45\textwidth]{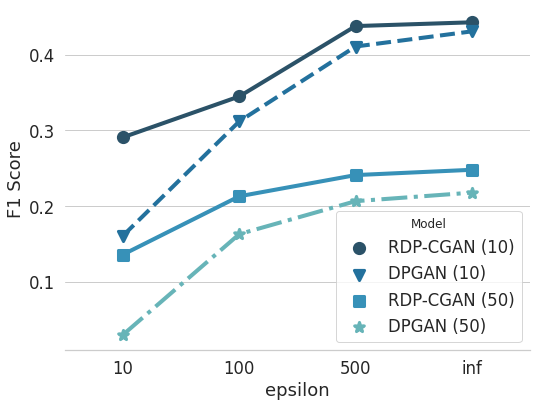}
\caption{The prediction accuracy of models trained on synthetic data by varying privacy budget. The number of top features used for training are `10' and `50'. The averaged F-1 score of all models to train on the \textit{real data} is \textit{0.44} and \textit{0.35} for picking top-10 and top-50 features respectively.}
\label{fig:dwpred}
\end{figure}

\textbf{Evaluation Metrics:} To assess the quality of synthetically generated data, we used two evaluation metrics: 

\begin{enumerate}[leftmargin=*]
    \item \textbf{Maximum Mean Discrepancy~(MMD):} This statistical measure indicates the extent to which the model captures the statistical distribution of the real data. In a recent study~\cite{xu2018empirical}, MMD demonstrated most of the desired features of an evaluation metric, especially for GANs. This is used in an \textit{unsupervised setting} since there are no labeled data for statistical measurements. To report MMD, we compared two samples of real and synthetic data, each of size 10000.
    \item \textbf{Dimension-wise prediction:} This setting aims to illustrate the inter-connection between features, i.e., if we can predict missing features using the features available in the dataset. We select top-10 and top-50 most frequent features~(ICD-9 codes in the patients' history) from training, test, and synthetic sets. In each run, one testing dimension~($k$) from $\mathcal{D}_{syn}$ and $\mathcal{D}_{tr}$ will be selected as $\mathcal{D}_{syn,k} \in \{0,1\}^{N \times 1}$ and $\mathcal{D}_{tr,k} \in \{0,1\}^{N \times 1}$. The remaining dimensions~($\mathcal{D}_{syn, \backslash k} \in \{0,1\}^{N \times 1}$ and $\mathcal{D}_{tr, \backslash k} \in \{0,1\}^{N \times 1}$) will be utilized to train a classifier that aims to predict $\mathcal{S}_{te,k} \in \{0,1\}^{N \times 1}$ from the test set.~We employed Random Forests~\cite{breiman2001random},~XGBoost~\cite{chen2016xgboost},~and Decision Tree~\cite{quinlan1986induction}.~We report the averaged performance over all predictive models~(trained on synthetic data) and for all features using the \textit{F1-score}~(Figure~\ref{fig:dwpred}). \textit{~We compare models by considering different privacy budgets by fixing $\delta=10^{-5}$ and varying $\epsilon$.}
\end{enumerate}

As can be observed in Figure~\ref{fig:mimicmmd} and Figure~\ref{fig:dwpred}, without enforcing privacy ($\epsilon$), our model performs better compared to DPGAN. \textit{This is due to the fact that by using 1-D convolutional architecture, our model captures the correlated information among adjacent ICD-9 codes in the MIMIC-III dataset. Top-features indicate the most frequent features in the database.}

To further investigate the effect of CGANs and CAEs in capturing correlated features, we conduct some experiments without enforcing any privacy and compare our method with existing methods. As can be observed in Table~\ref{tab:dwpredmimictopfeatures} our method outperforms other methods. Note that when the number of top features are increased, our method shows even better performance compared to other methods since capturing correlation for higher number of features becomes more feasible with convolutional layers.

    

\begin{table}
  \caption{Comparison of different methods using \textit{F-1 score} for the dimension-wise prediction setting. Except for the \textit{Real Data} column, the classifiers are trained on the synthetic data. The closer the results are to the Real Data experiments, the higher the quality of synthetic data is and hence the model is better.}
  \label{tab:dwpredmimictopfeatures}
  \centering
  \setlength{\tabcolsep}{3pt}
  \resizebox{.48\textwidth}{!}{
\begin{tabular}{c||c|c|c|c}
\toprule
    
    Top Features & Real Data & MedGAN & DPGAN & RDP-CGAN\\
    \midrule
    10 & 0.51 $\pm$ 0.043  &   0.41  $\pm$ 0.013  & 0.36 $\pm$ 0.010 & \textbf{0.44} $\pm$ \textbf{0.017}\\
    50 &  0.45 $\pm$ 0.037  &   0.24 $\pm$ 0.017  & 0.21 $\pm$ 0.017 & \textbf{0.35} $\pm$ \textbf{0.056}\\
    100 &  0.36 $\pm$ 0.051  &   0.15 $\pm$ 0.016  & 0.12 $\pm$ 0.009 & \textbf{0.25} $\pm$ \textbf{0.096}\\
    \bottomrule
\end{tabular}
}
\end{table}

\subsection{Supervised Synthetic Data Generation}

\begin{table*}
\normalsize
  \caption{Summary statistics of the datasets used in this paper along with the performance of the base model.}
  \label{tab:summarystats}
  \centering
  \setlength{\tabcolsep}{3pt} 

\begin{tabular}{c||c|c|c|c|c}
\toprule 

    Dataset & Samples & Features & Positive Labels & AUROC & AUPRC \\
    \midrule
    MIMIC-III & 12,127 & 31 & 3272~(27\%)  &    0.81 $\pm$ 0.005 & 0.74  $\pm$ 0.007\\
    Kaggle Cervical Cancer & 858 & 36  & 52~(6\%)  &   0.94 $\pm$ 0.011  & 0.69  $\pm$ 0.003\\
    UCI Epileptic Seizure & 11,500 & 178 & 2300~(20\%)  &   0.97   $\pm$ 0.008 & 0.94  $\pm$ 0.014\\
    PTB Diagnostic & 14,552 & 118 & 10506~(62\%)  &   0.97  $\pm$ 0.006 & 0.96  $\pm$ 0.003\\
    Kaggle Cardiovascular Disease & 70,000   &   14  & 35000~(50\%) &  0.80   $\pm$ 0.017 & 0.78  $\pm$ 0.011\\
    MIT-BIHArrhythmia & 109,444   &   188  & 18606~(17\%) &  0.94  $\pm$ 0.006 & 0.89  $\pm$ 0.014\\
    \bottomrule
\end{tabular}

\end{table*}

In this part, we consider a supervised setting and generate labeled synthetic data. For the experiments of this part, we need labeled data. The supervised setting here includes various classification tasks on different datasets.  In the experiments of this section, to pretrain the autoencoder, we used Eq.~\ref{eq:mse} as the loss function.

\textbf{Data Processing:} For all the datasets except for MIMIC, we follow the same data processing steps that are described in  Section~\ref{sub:datasets}. For the MIMIC-III dataset, we used one of the top three codes across hospital admissions~\cite{johnson2016mimic}, which is 414.01 (associated with \textit{coronary atherosclerosis}) and extracted patients diagnosed with that specific medical code. The classification task will be mortality prediction. We used 12,127 unique admissions and variables such as demographic information~(Marital status, ethnicity, and insurance status), admission
Information~(e.g., days of admission), treatment information~(cardiac defibrillator implant with/without cardiac catheterization), diagnostic information~(respiratory disorder, cancer, etc.), and lab results~(kidney function tests, creatine kinase, etc.) forming a total of 31 variables. Each admission is considered as one data sample. Since multiple admissions might be associated with one patient we will have multiple samples per patient. This comes from the fact that, in a 1-year interval, a patient may survive, but within the next few years and with new medical conditions, the same patient may not survive in the 1-year observation windows.

\textbf{Evaluation:} Let us consider the following two settings: \textit{(A)} Train and test the model on the real data as the baseline~(results shown in Table~\ref{tab:summarystats}). \textit{(B)} Train on the generated data and test on the real data (this setting is used to quantify the quality of synthetic data). \textit{Note that the class distribution of the generated synthetic data must be identical to the real data}. Setting (B) can demonstrate how well the model has performed the task of synthetic data generation.~\textit{The closer the performance of setting~(B) is to setting~(A), the better the model is in terms of the quality of the generated synthetic data}. We conducted different sets of experiments. We conduct ten runs (E=10), for each experiment, and report the averaged {AUROC}~(Area Under the ROC curve) and \textit{AUPRC}~(Area Under the Precision-Recall Curve) for the models' evaluations. The AUPRC metric is being utilized here since it operates better than AUROC for an imbalanced classification setting.

\begin{table*}
\normalsize
  \caption{Performance comparison of different methods using AUROC metric under no privacy constraints.}
  \label{tab:datasetsclassifierauroc}
  \centering
\begin{tabular}{c||c|c|c|c|c}
\toprule
    
    Dataset & MedGAN & TableGAN & DPGAN & PATE-GAN & RDP-CGAN\\
    \midrule
    MIMIC-III  & 0.71 $\pm$ 0.011 & \underline{0.72 $\pm$ 0.017} & 0.71 $\pm$  0.014 &   0.70 $\pm$  0.007 & \textbf{0.74} $\pm$ \textbf{0.012}\\
    Kaggle Cervical Cancer   & 0.89 $\pm$ 0.010 & \underline{0.90 $\pm$ 0.012} & 0.90  $\pm$ 0.016  &   0.89 $\pm$ 0.009 & \textbf{0.92} $\pm$ \textbf{0.009}\\
    UCI Epileptic Seizure   & 0.87 $\pm$ 0.013 & \textbf{0.91} $\pm$ \textbf{0.019} & 0.88 $\pm$  0.024 &   0.85 $\pm$ 0.014  & \underline{0.90 $\pm$ 0.011}\\
    PTB Diagnostic     & 0.86 $\pm$ 0.018 & \textbf{0.94} $\pm$ \textbf{0.009} &  0.88 $\pm$ 0.018 & 0.91 $\pm$ 0.006 & \underline{0.93 $\pm$ 0.016}\\
    Kaggle Cardiovascular Disease    & 0.71 $\pm$ 0.031 & \underline{0.73 $\pm$ 0.016} &   0.71 $\pm$ 0.019 & 0.72 $\pm$ 0.008 & \textbf{0.76} $\pm$ \textbf{0.014}\\
    MIT-BIHArrhythmia     & 0.90 $\pm$ 0.021  & \underline{0.89 $\pm$ 0.019}  &   0.88 $\pm$ 0.012 & 0.86 $\pm$ 0.014 & \textbf{0.90} $\pm$ \textbf{0.009}\\
    \bottomrule
\end{tabular}
\end{table*}

\begin{table*}
\normalsize
  \caption{Performance comparison of different methods using AUPRC metric under no privacy constraints.}
  \label{tab:datasetsclassifierauprc}
  \centering
\begin{tabular}{c||c|c|c|c|c}
\toprule
    
    Dataset& MedGAN & TableGAN & DPGAN & PATE-GAN& RDP-CGAN \\
    \midrule
    MIMIC-III & 0.68 $\pm$ 0.011 & \underline{0.71 $\pm$ 0.007} & 0.69 $\pm$ 0.006 & 0.70 $\pm$ 0.019  & \textbf{0.72} $\pm$ \textbf{0.009}\\
    Kaggle Cervical Cancer& 0.55 $\pm$ 0.012 & \underline{0.60 $\pm$ 0.019}  & 0.59 $\pm$ 0.008 &   0.58 $\pm$ 0.020 & \textbf{0.62} $\pm$ \textbf{0.017} \\
    UCI Epileptic Seizure & 0.79 $\pm$ 0.013 & \textbf{0.86} $\pm$ \textbf{0.016} & 0.82 $\pm$ 0.015  &   0.81 $\pm$ 0.012  & \underline{0.84 $\pm$ 0.020}\\
    PTB Diagnostic & 0.88 $\pm$ 0.006  & \textbf{0.93} $\pm$ \textbf{0.008} & 0.90 $\pm$ 0.007    &   0.88 $\pm$ 0.014 & \underline{0.92 $\pm$ 0.012}\\
    Kaggle Cardiovascular Disease & 0.69 $\pm$ 0.021 & \underline{0.73 $\pm$ 0.011} & 0.73 $\pm$ 0.021  &   0.72 $\pm$ 0.017  & \textbf{0.75} $\pm$ \textbf{0.009}\\
    MIT-BIHArrhythmia  & 0.77 $\pm$ 0.016  & \underline{0.85 $\pm$ 0.013} & 0.82 $\pm$ 0.008 &   0.81 $\pm$  0.005 & \textbf{0.86} $\pm$ \textbf{0.013}\\
    \bottomrule
\end{tabular}
\end{table*}

\subsubsection{The effect of architecture}

In order to investigate the effect of convolutional GANs and convolutional autoencoders in our model, we perform some experiments with \textbf{\textit{no privacy enforcement}}. The case $\epsilon=\infty$ corresponds to the setup where we do not enforce any privacy. The AUROC and AUPRC results are reported in Table~\ref{tab:datasetsclassifierauroc} and Table~\ref{tab:datasetsclassifierauprc}, respectively. In the aforementioned tables, the best and second best results are indicated with \textbf{bold} and \underline{underline} text.  The classifiers are trained on the synthetic data. The closer the results are to the Real Data experiments, the higher the quality of synthetic data is and hence, we have a better model. As can be observed from those tables, for the challenging datasets where there is a mixture of continuous and  discrete variables, our model outperforms others since it not only captures feature correlations but also utilizes convolutional autoencoders which allow our approach learn robust representations for the mixture of data types. For other datasets, TableGAN closely competes with our model. This is because TableGAN also uses convolutional GANs as well as auxiliary classifiers~\cite{odena2017conditional} for improving the performance of the GANs. We could incorporate auxiliary classifiers to improve our model, however, it would narrow down the scope of our work since using such auxiliary models would make it infeasible for unsupervised synthetic data generation.

\subsubsection{The effect of differential privacy}

\begin{table*}
  \caption{The comparison of different models under the $(1,10^{-5})$-DP setting.~For $\epsilon=\infty$, we used our RDP-CGAN model without enforcing privacy. We used synthetic data for training all models. The best and second best results are indicated with \textbf{bold} and \underline{underline} text.}
  \label{tab:modelsuperviseddp}
  \centering
  \setlength{\tabcolsep}{3.5pt}
  \resizebox{.99\textwidth}{!}{
\begin{tabular}{c||c|c|c|c||c|c|c|c}
\toprule 
      & \multicolumn{4}{c}{AUROC} & \multicolumn{4}{c}{AUPRC}\\
    \cmidrule(r){2-9}
    
    Dataset & $\epsilon=\infty$ & DPGAN & PATE-GAN & RDP-CGAN
    & $\epsilon=\infty$ & DPGAN & PATE-GAN & RDP-CGAN  \\
    \midrule
    MIMIC-III & 0.74 $\pm$ 0.012   & 0.59 $\pm$ 0.009    &   \underline{0.62 $\pm$ 0.011}  & \textbf{0.67} $\pm$ \textbf{0.010} & 0.72 $\pm$ 0.009 & 0.55 $\pm$ 0.013 & \underline{0.58 $\pm$ 0.017} & \textbf{0.62} $\pm$ \textbf{0.015} \\
    Kaggle Cervical Cancer & 0.92 $\pm$ 0.009 & 0.86 $\pm$ 0.009    &   \textbf{0.91} $\pm$ \textbf{0.005}  & \underline{0.89 $\pm$ 0.005} & 0.62 $\pm$ 0.017 & 0.53 $\pm$ 0.008 & \underline{0.54 $\pm$ 0.014} & \textbf{0.57} $\pm$ \textbf{0.007} \\
    UCI Epileptic Seizure & 0.90 $\pm$ 0.011   & 0.72 $\pm$ 0.009    &   \underline{0.74 $\pm$ 0.014}  & \textbf{0.84} $\pm$ \textbf{0.008} & 0.84 $\pm$ 0.020 & 0.57 $\pm$ 0.003 & \underline{0.63 $\pm$ 0.016} & \textbf{0.69} $\pm$ \textbf{0.019}\\
    PTB Diagnostic ECG & 0.93 $\pm$ 0.016   & 0.71 $\pm$ 0.012    &   \underline{0.75 $\pm$ 0.012}  & \textbf{0.79} $\pm$ \textbf{0.009} & 0.92 $\pm$ 0.012 & 0.71 $\pm$ 0.018 & \underline{0.76 $\pm$ 0.011} & \textbf{0.80} $\pm$ \textbf{0.008}\\
    Kaggle  Cardiovascular & 0.76 $\pm$ 0.014   & 0.61 $\pm$ 0.019    &   \underline{0.66 $\pm$ 0.006}  & \textbf{0.69} $\pm$ \textbf{0.013} & 0.75 $\pm$ 0.009 & 0.60 $\pm$ 0.001 & \underline{0.63 $\pm$ 0.007} & \textbf{0.66} $\pm$ \textbf{0.016}\\
    MIT-BIHArrhythmia & 0.90 $\pm$ 0.009 & 0.69 $\pm$ 0.004  &   \underline{0.73 $\pm$ 0.006}  & \textbf{0.77} $\pm$ \textbf{0.003} & 0.86 $\pm$ 0.013 & 0.68 $\pm$ 0.023 & \underline{0.73 $\pm$ 0.016} & \textbf{0.78} $\pm$ \textbf{0.008}\\
    \bottomrule
\end{tabular}
}
\end{table*}

 We will now investigate how well different methods perform in the differential privacy setting.~\textit{The base model is when we train using setting (A), which will have the highest accuracy as expected}. So the question we investigate here is: \textit{how much will the accuracy drop for different models at the same level of privacy budget~$(\epsilon,\delta)$?} We compare our model with privacy-preserving models~(see Table~\ref{tab:methodscomparisoncriteria}). Table \ref{tab:modelsuperviseddp} shows these results for the supervised setting described above. In the majority of the experiments, the synthetic data generated by our model, demonstrates higher quality for classification tasks compared to other models \textit{under the same privacy budget}.


\subsubsection{The effect of privacy budget}
\begin{figure*}[t!]
    \centering
    \begin{subfigure}[t]{0.30\textwidth}
        \centering
        \includegraphics[height=1.75in]{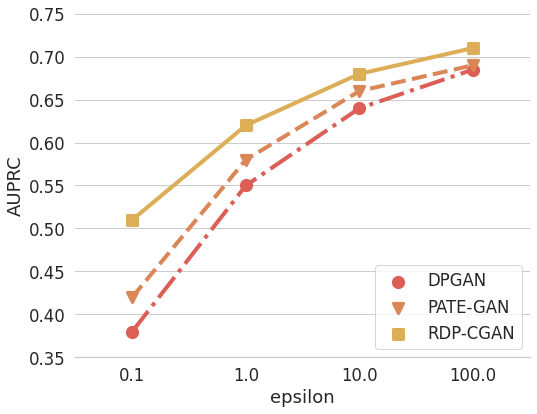}
        \caption{MIMIC-III.}
    \end{subfigure}
    \begin{subfigure}[t]{0.30\textwidth}
        \centering
        \includegraphics[height=1.75in]{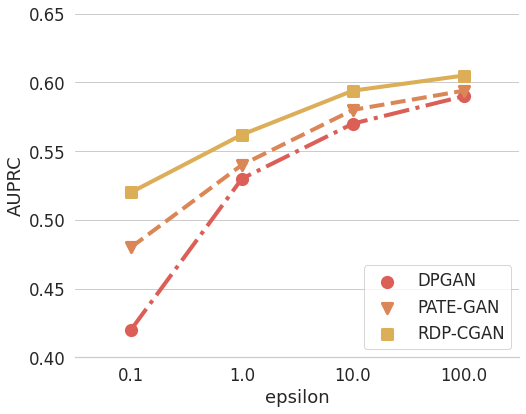}
        \caption{Kaggle Cervical Cancer.}
    \end{subfigure}
    \begin{subfigure}[t]{0.30\textwidth}
        \centering
        \includegraphics[height=1.75in]{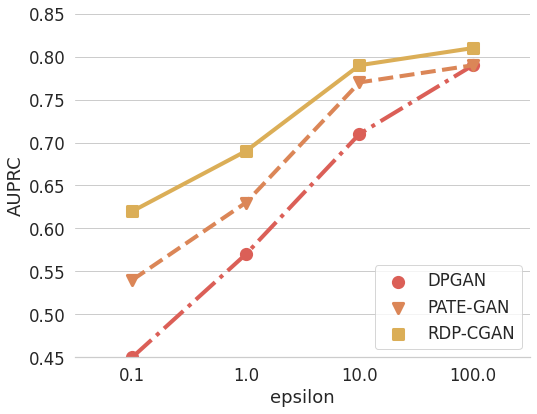}
        \caption{UCI Epileptic Seizure.}
    \end{subfigure}
    \begin{subfigure}[t]{0.30\textwidth}
        \centering
        \includegraphics[height=1.75in]{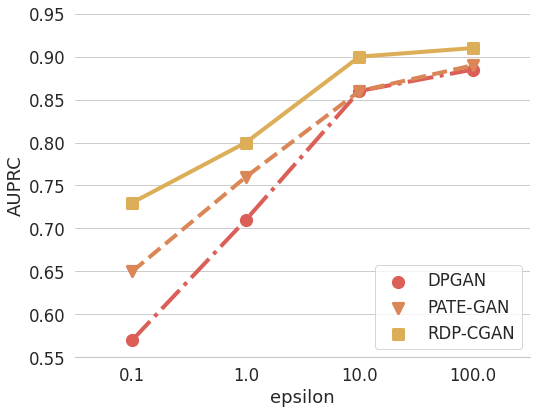}
        \caption{PTB Diagnostic.}
    \end{subfigure}
    \begin{subfigure}[t]{0.30\textwidth}
        \centering
        \includegraphics[height=1.75in]{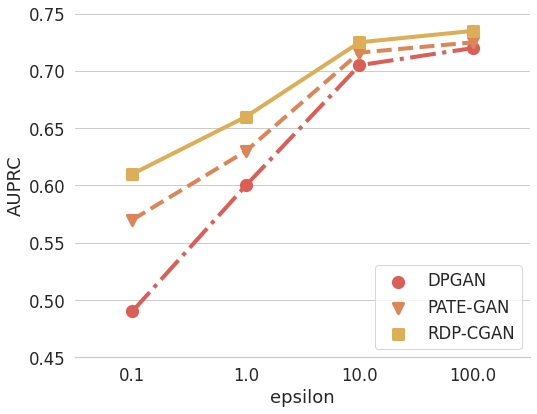}
        \caption{Kaggle Cardiovascular.}
    \end{subfigure}
    \begin{subfigure}[t]{0.30\textwidth}
        \centering
        \includegraphics[height=1.75in]{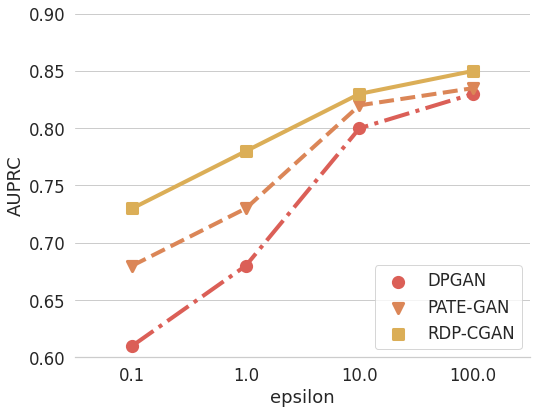}
        \caption{MIT-BIH Arrhythmia.}
    \end{subfigure}
    \caption{The effect of privacy budget on the synthetic data quality measured by AUPRC. The baseline is associated with $\epsilon=\infty$ for which we trained each model with no privacy constraint (see Table~\ref{tab:datasetsclassifierauprc}). Higher AUPRC is associated with higher quality of synthetic data and hence represents a better model.} 
    \label{fig:privacybudget}
\end{figure*}

We also investigate how models will perform under different privacy budgets. Fig.~\ref{fig:privacybudget} demonstrates the trade-off between the privacy budget and the synthetic data quality. RDP-CGAN consistently shows better performance compared to other benchmarks in all the datasets. Our approach demonstrates it is particularly effective in lower privacy budgets. As an example, for the Kaggle Cervical Cancer dataset and for $\epsilon={100,10,1,0.1}$, our model achieves significantly higher AUPRC compared to the PATE-GAN.

\subsubsection{Ablation study}
We will now conduct an ablation study to investigate the effect of each component of our model. In particular, we are interested in assessing the importance and impact of utilizing autoencoders and convolutional architectures. The results of our ablation study are reported in Table~\ref{tab:ablation-AUPRC} and Table~\ref{tab:ablation-AUROC}. The various scenarios reported are described below:

{\begin{itemize}[leftmargin=*]
    \item \textit{W/O CAE:} without convolutional architecture in autoencoder.
    \item \textit{W/O AE:} without autoencoder.
    \item \textit{W/O CG:} without convolutional architecture in GAN's generator.
    \item \textit{W/O CD:} without convolutional architecture in GAN's discriminator.
    \item \textit{W/O CDCG:} a standard GAN architecture with MLP.
\end{itemize}}

{The results shown in Table~\ref{tab:ablation-AUPRC} and Table~\ref{tab:ablation-AUROC} demonstrate the importance of convolutional architectures. The interesting finding here is that, for the datasets with only continuous values like UCI Epileptic Seizure, PTB Diagnostic ECG, and MIT-BIHArrhythmia) (as opposed to a mixture of continuous-discrete variables), the use of autoencoders actually downgraded the system performance. Hence, we can conclude that autoencoders are very useful in the presence of discrete data or a mixture of both continuous and discrete data types.}

\begin{table*}
  \caption{Ablation study results. For the baseline $\epsilon=\infty$, we used our RDP-CGAN model without enforcing privacy. Each column corresponds to a change in the core RDP-CGAN architecture. The reported results are AUROC.}
  \label{tab:ablation-AUROC}
  \centering
  \setlength{\tabcolsep}{3.5pt}
  \resizebox{.9\textwidth}{!}{
\begin{tabular}{c||c|c|c|c|c|c}
\toprule 
    
    Dataset & $\epsilon=\infty$ & W/O CAE & W/O AE
    & W/O CG & W/O CD & W/O CDCG  \\
    \midrule
    MIMIC-III & 0.74 $\pm$ 0.012   & 0.73 $\pm$ 0.017   & 0.69 $\pm$ 0.031 & 0.72 $\pm$ 0.023 & 0.71 $\pm$ 0.027 & 0.70 $\pm$ 0.016 \\
    Kaggle Cervical Cancer & 0.92 $\pm$ 0.009 & 0.90 $\pm$ 0.019    & 0.86 $\pm$ 0.027 & 0.92 $\pm$ 0.031 & 0.90 $\pm$ 0.014 & 0.88 $\pm$ 0.021 \\
    UCI Epileptic Seizure & 0.90 $\pm$ 0.011 & 0.89 $\pm$ 0.016 & 0.91 $\pm$ 0.021 & 0.88 $\pm$ 0.022 & 0.86 $\pm$ 0.008 & 0.85 $\pm$ 0.026 \\
    PTB Diagnostic ECG & 0.93 $\pm$ 0.016   & 0.91 $\pm$ 0.011 & 0.94 $\pm$ 0.015 &  0.92 $\pm$ 0.029 & 0.91 $\pm$ 0.027 & 0.89 $\pm$ 0.014 \\
    Kaggle  Cardiovascular & 0.76 $\pm$ 0.014 & 0.74 $\pm$ 0.012    & 0.71 $\pm$ 0.041 & 0.43 $\pm$ 0.015 & 0.72 $\pm$ 0.021 & 0.71 $\pm$ 0.026 \\
    MIT-BIHArrhythmia & 0.90 $\pm$ 0.009  & 0.88 $\pm$ 0.013   & 0.90 $\pm$ 0.014 & 0.88 $\pm$ 0.012 & 0.86 $\pm$ 0.017 & 0.84 $\pm$ 0.041 \\
    \bottomrule
\end{tabular}
}
\end{table*}

\begin{table*}
  \caption{Ablation study results. For the baseline $\epsilon=\infty$, we used our RDP-CGAN model without enforcing privacy. Each column corresponds to a change in the core RDP-CGAN architecture. The reported results are AUPRC.}
  \label{tab:ablation-AUPRC}
  \centering
  \setlength{\tabcolsep}{3.5pt}
  \resizebox{.9\textwidth}{!}{
\begin{tabular}{c||c|c|c|c|c|c}
\toprule 
    
    Dataset & $\epsilon=\infty$ & W/O CAE & W/O AE
    & W/O CG & W/O CD & W/O CDCG  \\
    \midrule
    MIMIC-III & 0.72\ $\pm$ 0.009   & 0.70\ $\pm$ 0.032   & 0.67 $\pm$ 0.054 & 0.70 $\pm$ 0.076 & 0.69 $\pm$ 0.023 & 0.66 $\pm$ 0.065 \\
    Kaggle Cervical Cancer & 0.62 $\pm$ 0.017 & 0.61 $\pm$ 0.031    & 0.58 $\pm$ 0.017 & 0.59 $\pm$ 0.036 & 0.59 $\pm$ 0.041 & 0.55 $\pm$ 0.023 \\
    UCI Epileptic Seizure & 0.84 $\pm$ 0.020  & 0.82 $\pm$ 0.034    & 0.86 $\pm$ 0.041 & 0.79 $\pm$ 0.024 & 0.81 $\pm$ 0.012 & 0.80 $\pm$ 0.018 \\
    PTB Diagnostic ECG & 0.92 $\pm$ 0.012   & 0.90 $\pm$ 0.028 & 0.94 $\pm$ 0.031 &  0.89 $\pm$ 0.022 & 0.88 $\pm$ 0.022 & 0.88 $\pm$ 0.029 \\
    Kaggle  Cardiovascular & 0.75 $\pm$ 0.009 & 0.73 $\pm$ 0.052    & 0.71 $\pm$ 0.017 & 0.72 $\pm$ 0.033 & 0.71 $\pm$ 0.041 & 0.70 $\pm$ 0.014 \\
    MIT-BIHArrhythmia & 0.86 $\pm$ 0.013  & 0.83 $\pm$ 0.019    & 0.87 $\pm$ 0.026 & 0.85 $\pm$ 0.038 & 0.84 $\pm$ 0.023 & 0.82 $\pm$ 0.032 \\
    \bottomrule
\end{tabular}
}
\end{table*}

\section{Conclusion}\label{sec:conclusions}

In this work, we proposed and developed a differentially private framework for synthetic data generation using Rényi Differential Privacy. The model aimed to capture the temporal information and feature correlation using convolutional neural networks. We empirically demonstrate that by employing convolutional autoencoders we can effectively handle variables that are continuous, discrete or a mixture of both.~We argue that we will need to secure both encoder and decoder parts of the autoencoder since there is no guarantee that the autoencoder's training is being done by a trusted third-party. We show our model outperforms other models under the same privacy budget. This phenomenon may come in part from reporting a tighter bound and, in part, from utilizing convolutional networks. We reported the performance of different models by replacing real data with synthetic data for training our machine learning models.

\section*{Acknowledgements}
This work was supported in part by the US National Science Foundation grants IIS-1619028, IIS-1707498, IIS-1838730 and NVIDIA Corp. We also acknowledge the support provided by NewWave Telecom Technologies during the early stages of this project through a fellowship award.


%





\ifCLASSOPTIONcaptionsoff
  \newpage
\fi



%
\bibliography{main}
\bibliographystyle{ieeetr}

%


\begin{IEEEbiography}[{\includegraphics[width=1in,clip]{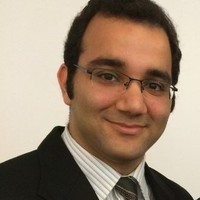}}]{Amirsina Torfi} holds a Ph.D. in Computer Science from Virginia Tech, and a B.S. and M.S. from Iran University of Science and Technology in Electrical Engineering and Information Technology, respectively. His research interests include Machine Learning and Deep Learning in various applications such as NLP, healthcare, and computer vision. He published several peer-reviewed. He is the founder of Instill AI, a start-up company aimed to enhance the utilization of AI in real-world applications. He is also a leading expert in open-source development as some of his Deep Learning projects are well-known worldwide.

\end{IEEEbiography}

\begin{IEEEbiography}[{\includegraphics[width=1in,clip]{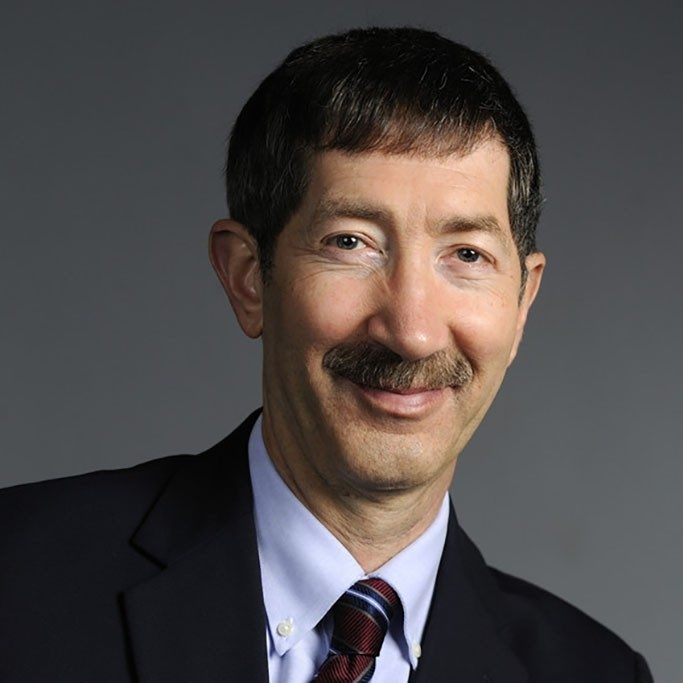}}]{Edward A.~Fox} holds a Ph.D. and M.S. in Computer Science from Cornell University, and a B.S. from M.I.T. He is a Fellow of both ACM and IEEE, cited for leadership in digital libraries and information retrieval. Since 1983 he has been at Virginia Polytechnic Institute and State University (VPI\&SU or Virginia Tech), where he serves as Professor of Computer Science, and by courtesy, of ECE. He was an elected member of the Board of Directors of the Computing Research Association and was Chair (now a member) of the ACM/IEEE-CS Joint Conference on Digital Libraries (JCDL) Steering Committee. He served on the IEEE Thesaurus Editorial Board, and was Chairman of the IEEE-CS Technical Committee on Digital Libraries. He works in the areas of digital libraries, information storage and retrieval, machine learning/AI, computational linguistics (NLP), hypertext/hypermedia/multimedia, computing education, CD-ROM and optical disc technology, electronic publishing, and expert systems. In these areas, he has participated in, organized, presented, and/or reviewed for hundreds of conferences and workshops. 131 grants have funded his research. For the Association for Computing Machinery he served 2018-2019 as a member of its Publications Board (and as co-chair of its digital libraries committee). He was editor for information retrieval and digital libraries in the ACM Books series. He was founder and co-editor-in-chief for the ACM Journal of Educational Resources in Computing, was a member of the editorial board for ACM Transactions on Information Systems, and was General Chair for JCDL 2001. He served as Program Chair for ACM DL'99, ACM DL'96, and ACM SIGIR'95 - and Program Co-chair for JCDL 2018, CIKM 2006, and ICADL 2005. He serves on the editorial boards of Journal of Educational Multimedia and Hypermedia, Journal of Universal Computer Science, and PeerJ Computer Science.
\end{IEEEbiography}

\begin{IEEEbiography}[{\includegraphics[width=1in,clip]{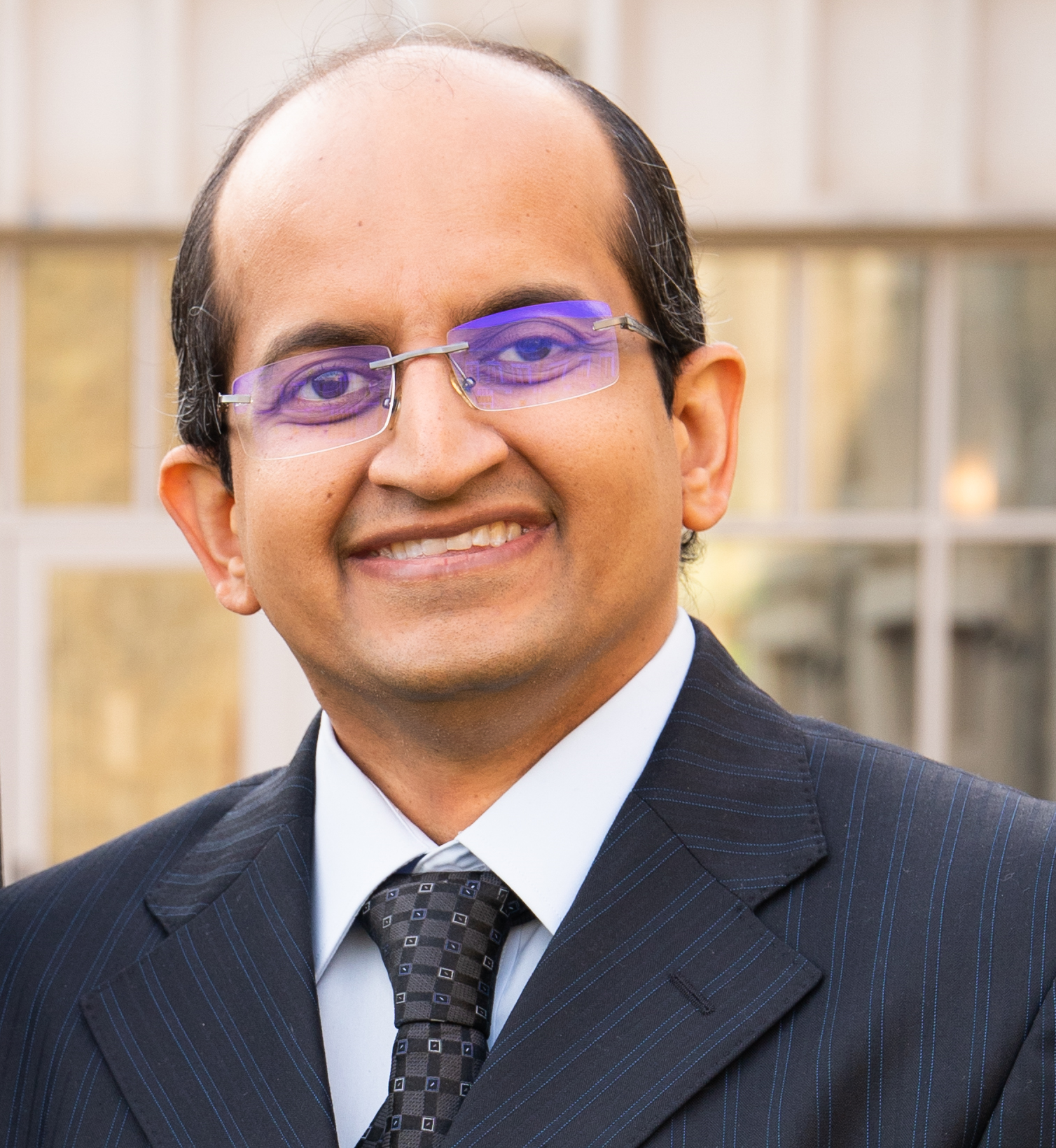}}]{Chandan K. Reddy}
is a Professor in the Department of Computer Science at Virginia Tech. He received his Ph.D. from Cornell University and M.S. from Michigan State University. His primary research interests are Data Mining and Machine Learning with applications to various real-world domains including healthcare, transportation, social networks, and e-commerce. He has published over 130 peer-reviewed articles in leading conferences and journals. He received several awards for his research work including the Best Application Paper Award at ACM SIGKDD conference in 2010, Best Poster Award at IEEE VAST conference in 2014, Best Student Paper Award at IEEE ICDM conference in 2016, and was a finalist of the INFORMS Franz Edelman Award Competition in 2011.  He is serving on the editorial boards of ACM TKDD, IEEE Big Data, and DMKD journals. He is a senior member of the IEEE and distinguished member of the ACM.
\end{IEEEbiography}





\end{document}